# On the Adaptability of Neural Network Image Super-Resolution

Kah Keong Chua, Yong Haur Tay

Centre for Computing and Intelligent System

Universiti Tunku Abdul Rahman
Kuala Lumpur, Malaysia
`chua.kah.keong@gmail.com`
`tayyh@utar.edu.my`

*Abstract*. In this paper, we described and developed a framework for Multilayer Perceptron (MLP) to work on low level image processing, where MLP will be used to perform image super-resolution. Meanwhile, MLP are trained with different types of images from various categories, hence analyse the behaviour and performance of the neural network. The tests are carried out using qualitative test, in which Mean Squared Error (MSE), Peak Signal-to-Noise Ratio (PSNR) and Structural Similarity Index (SSIM). The results showed that MLP trained with single image category can perform reasonably well compared to methods proposed by other researchers.

*Keywords*: **Image Super Resolution, Neural Network, Multilayer Perceptron, Mean Squared Error, Peak Signal-to-Noise Ratio, Structural Similarity Index.**

## 1      Introduction

In this modern era of technology, a modern image sensor is usually either a charge-coupled device (CCD), or a complementary metal-oxide-semiconductor (CMOS) active-pixel sensor. In order to increase the resolution of the image captured, the size of each individual pixel sensor must be reduced to obtain higher sensor density, which is a challenge in terms of the cost of fabrication. Hence, a post-processing technique is introduced to tackle the image degradation problem, which commonly referred as super resolution reconstruction. Super-Resolution, generally speaking, is the process of recovering a high-resolution (HR) image from single or multiple low-resolution (LR) images.

A variety of approaches for solving super-resolution problem have been proposed. Foremost attempts worked in the frequency domain by Tsai and Huang [3], regularly recovering higher frequency components by taking the advantage of shifting and aliasing properties of the Fourier transform. In order to overcome the image artefacts caused by commonly used methods, such as bilinear or bicubic interpolation, which includes jagged edges and blurred contours, many new methods and algorithms have been proposed to upscale the images without introducing much artefacts [4-6].

Interpolation-based approach, on the other hand, reconstructs a high-resolution (HR) image by projecting the low-resolution (LR) images obtained to the respective referencing image, then follows by blending all the accessible information from each image to form the final result [4]. An interpolation-based up-scaling method was proposed based on a two-step grid filling and iterative correction of the interpolated pixels [2], which perform quite well with ability to minimize artefacts without introducing too artificial details.

On the other hand, neural network, or known as learning-based techniques, is commonly used in image super-resolution. Neural network can be trained to estimate nonlinear functions as well as perform parallel computation, hence they a potential means of minimizing the processing time of super-resolution systems [5]. Most of the learning-based super-resolution works in analysing and processing huge datasets from a particular image class, for instance faces or alphabets [6]. The images are mapped from low resolution (LR) to high resolution (HR) images through the process.

In fact, learning-based method requires large database and this consumes considerable amount of time during the training process. Throughout the process, the neural network is trained while weight factors are adjusted and fine-tuned every epoch [6]. Most of the results obtained from learning-based methods are closely relates to the amount of training samples provided, as the trained results are approximated from non-identical images [10-11].

Over the years, there are many algorithms and methods proposed by various researchers, yet neural network based image up-scaling methods gain their popularity due to the flexibility and learning capability of the network. The nature of a neural network which allows fine tuning after the training completed greatly improve the adaptation of the network, hence increase the performance of it.

In a nutshell, the contribution of our work in this paper is to perform low level image processing using neural network, which is image super resolution, by creating a framework to test the behaviour of neural network based on different types of training samples provided, in contrast with several other methods proposed by other researchers in image super-resolution.

This paper is organized as follows: Section II briefly describes the Multilayer Perceptron and its architecture, while Section III explains the creation of training and testing datasets. Furthermore, Section IV describes the experimental framework and the simulation results. Discussion is arranged in Section V, and it's concluded in Section VI alongside with future works. Acknowledgement and references are listed at the last part of the paper.

## 2  Multilayer Perceptrons

Multilayer Perceptron (MLP) is a feedforward backpropagation artificial neural network model which effectively maps input datasets to their respective output set. MLP works on a supervised learning technique namely backpropagation during training phase of the network.

MLP generally consist of three different layers, which are input layer, hidden layer and the output layer. Typically the entire neural network has an input layer and an output layer respectively, while the amount of hidden layers highly depends on the design of network. Each layer has multiple numbers of neurons, which varies according to the complexity of the network. The figure below shows a basic Multilayer Perceptron Neural Network model described by Phillip H. Sherrod [7]:

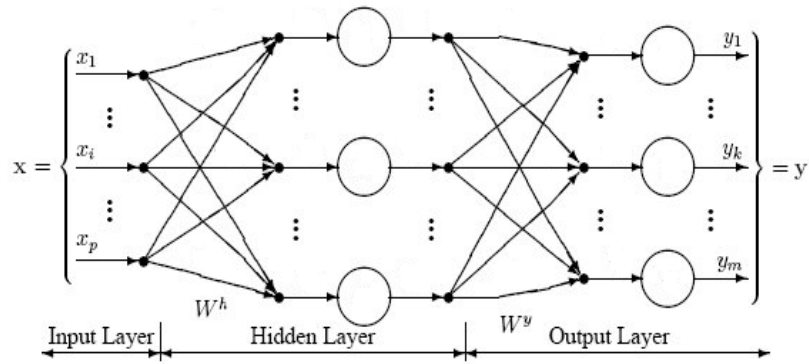

**Fig. 1.** A basic Multilayer Perceptron model

The input layer which consist variables of vector or matrix, is usually normalised in such that the range of each variable is from 0 to 1, while each variable is assigned accordingly to the neurons in hidden layer. Meanwhile, a constant input known as bias is multiplied by weight before passing to the hidden layer.

In the hidden layer, the neurons obtain the input from previous layer and multiply by a weight, before adding up to generate a weighted sum. Each output from the hidden layer is then distributed to the output layer. The values of each neuron in output layer will be multiplied by a weight factor before summing up and pass to a transfer function, which yields the output of the network.

The learning process takes place in the perceptron by altering the weight factors after each data samples are processed, or commonly known as training epoch. Weight factors are being adjusted accordingly by calculating the mean error of the expected result in contrary with the output result.

The aim of training a neural network is to search for a set of weight factors which link the provided target output with trained output as similar as it can get. Hence, the process of optimizing the number of hidden layers used and neuron amount in each layer greatly affects the performance of the entire network. In addition, the neural network ought to be validated throughout the training process to avoid over fitting.

## 3     System Overview

The MLP used for this research has been configured to match the requirement of the network. The MLP is a feedforward artificial neural network which consists of input layer, hidden layer and output layer. It is a typical one hidden layer MLP of 20 hidden units which has hyperbolic tangent (*tanh*) activation function, while the intermediate layer is of sigmoid function. The network takes in the desired pixel alongside with eight neighbouring pixel as reference, then outputs four up-scaled pixels. The process is repeated on the adjacent pixel, throughout the entire image. The figure below shows the architecture of the proposed MLP system:

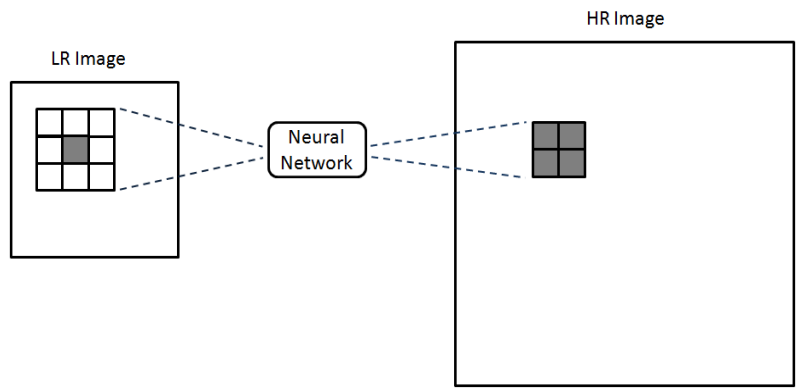

**Fig. 2.**   Architecture of the proposed MLP system

## 4     Creation of training and testing sets

For the datasets creation, we obtained the training images from morgueFile online archive (http://www.morguefile.com/) and all the images are subjected to the licence agreement available at their respective morgueFile webpage (http://www.morguefile.com/license/morguefile/). Meanwhile, we chose the images from different categories and background settings, for instance flowers, buildings, animals and cuisine as these sample images provide wide range of texture combination.

In addition, methods tested in this paper other than Multilayer Perceptron proposed above, are obtained then regenerated using the evaluation scripts and experimental setup available at the website: http://www.andreagiachetti.it/icbi

The images downloaded were then down-sampled to 256x256 pixels and 512x512 pixels, which correspond to low resolution (LR) and high resolution (HR) images respectively. All the files are RGB colour images and samples are extracted from all three colour channels. Moreover, a 3x3 kernel was used to sample the training sets from the red, green and blue channel of the images respectively.

From all the dataset generated, it is then further divided into 3 different categories, namely training set, validate set and testing set, at the ratio of 6:2:2. In each respective set, the samples are shuffled in order to avoid neural network from memorizing the input pattern.

The algorithms tested here include Nearest Neighbour, bicubic interpolation, an edge-oriented interpolation introduced by Chen et.al. [8], an improved version of new edge-directed interpolation (NEDI) [1], as well as the FBCI [2] and ICBI [2] proposed by Giachetti and Asuni, in addition with the Multilayer Perceptron based Neural Network described here.

## 5      Experiment Framework and Result

For the testing framework, all the methods will undergo a 2x image upscale process, in this case they are enlarged from 256x256 pixels to 512x512 pixels. The upscaled image will then compared to the original high resolution image to yield quantitative results. The difference of the images will be measured using Mean Square Error (MSE), Peak Signal-to-Noise Ratio (PSNR) as well as Structural Similarity Index (SSIM). The formulas are defined as follows.

Mean Squared Error is obtained by averaging the squared difference of a reference image and distorted image. The process takes place by summing up the squared difference of all the pixels, follows by dividing with the total pixels.

For images $A = \{a_1 \ldots a_M\}$, $B = \{b_1 \ldots b_M\}$, where M represents the number of pixels:

$$MSE(A, B) = \frac{1}{M} \sum_{i=1}^{M} (a_i - b_i)^2$$

(1)

Peak Signal-to-Noise Ratio is the ratio between the reference signal and the distorted signal for an image. Generally, higher PSNR value represents higher similarity of the reference image and distorted image.

For images $A = \{a_1 \ldots a_M\}$, $B = \{b_1 \ldots b_M\}$, where MAX is the maximum possible pixel value:

$$PSNR(A, B) = 10 \log_{10}(\frac{MAX^2}{MSE(A, B)})$$

(2)

Structural Similarity is calculated according to the change in luminance, contrast and structure of an image. Luminance is the average pixel intensity, the contrast is the variance between the reference and distorted image, while structure is obtained by cross-correlation the two images. Finally the SSIM value is calculated by averaging the resulting values.

Meanwhile, MLP will be trained using different sets of samples to tackle the problem effectively. The first training set will consist of sample from multiple sources, varies from flowers, buildings, animals, human and cuisine. This training set is customised at such to provide the neural network a wide range of samples, which will yield a more robust MLP network that will not suffer performance loss when comes to different setting.

Moreover, several other training sets are created solely on image samples from similar categories, in which different flowers will be sampled for flower training set, so on and so forth. MLP_gen is the neural network trained with samples obtained from multiple sources, while MLP_sp solely trained with the samples from same category.

**Table 1.** Test results on zebra

| Methods | Animal: Zebra | | |
|---|---|---|---|
| | **MSE** | **PSNR** | **SSIM** |
| N. Neighbour | 42.49 | 31.85 | 0.70 |
| Bicubic | 32.08 | 33.07 | 0.77 |
| Chen et.al. | 42.44 | 31.85 | 0.71 |
| NEDI | 41.07 | 32.00 | 0.72 |
| FCBI | 43.41 | 31.76 | 0.71 |
| ICBI | 40.65 | 32.04 | 0.73 |
| MLP_gen | 18.85 | 35.38 | 0.87 |
| MLP_sp | **16.62** | **35.92** | **0.90** |

**Table 2.** Test results on tiger

| Methods | Animal: Tiger | | |
|---|---|---|---|
| | **MSE** | **PSNR** | **SSIM** |
| N. Neighbour | 37.58 | 32.38 | 0.61 |
| Bicubic | 32.58 | 33.00 | 0.67 |
| Chen et.al. | 38.23 | 32.31 | 0.60 |
| NEDI | 37.81 | 32.35 | 0.61 |
| FCBI | 38.08 | 32.32 | 0.61 |
| ICBI | 36.70 | 32.48 | 0.63 |
| MLP_gen | 23.62 | 34.40 | **0.84** |
| MLP_sp | **22.46** | **34.62** | 0.83 |

**Table 3.** Test results on buildings

| Methods | Buildings | | |
|---|---|---|---|
| | **MSE** | **PSNR** | **SSIM** |
| N. Neighbour | 30.72 | 33.26 | 0.74 |
| Bicubic | 26.91 | 33.83 | 0.78 |
| Chen et.al. | 30.93 | 33.23 | 0.74 |
| NEDI | 30.76 | 33.25 | 0.74 |
| FCBI | 32.80 | 32.97 | 0.74 |
| ICBI | 31.37 | 33.17 | 0.76 |
| MLP_gen | **19.20** | **35.30** | **0.84** |
| MLP_sp | 22.83 | 34.55 | 0.72 |

**Table 4.** Test results on flowers

| Methods | Flowers | | |
|---|---|---|---|
| | MSE | PSNR | SSIM |
| N. Neighbour | 25.78 | 34.02 | 0.81 |
| Bicubic | 20.61 | 34.99 | 0.88 |
| Chen et.al. | 25.69 | 34.03 | 0.83 |
| NEDI | 24.83 | 34.18 | 0.84 |
| FCBI | 27.09 | 33.80 | 0.84 |
| ICBI | 24.90 | 34.17 | 0.86 |
| MLP_gen | 15.52 | 36.22 | 0.90 |
| MLP_sp | **11.90** | **37.38** | **0.92** |

**Table 5.** Test results on random texture

| Methods | Random Mixed Texture | | |
|---|---|---|---|
| | MSE | PSNR | SSIM |
| N. Neighbour | 60.22 | 30.33 | 0.58 |
| Bicubic | 52.81 | 30.90 | 0.62 |
| Chen et.al. | 61.06 | 30.27 | 0.56 |
| NEDI | 61.15 | 30.27 | 0.55 |
| FCBI | 62.16 | 30.20 | 0.56 |
| ICBI | 59.83 | 30.36 | 0.59 |
| MLP_gen | **31.06** | **33.21** | **0.84** |
| MLP_sp | 31.96 | 33.09 | **0.84** |

**Table 6.** Test results on colour vases

| Methods | Mixed Colour Vases | | |
|---|---|---|---|
| | MSE | PSNR | SSIM |
| N. Neighbour | 32.85 | 32.97 | 0.71 |
| Bicubic | 27.69 | 33.71 | 0.78 |
| Chen et.al. | 33.66 | 32.86 | 0.73 |
| NEDI | 34.21 | 32.79 | 0.72 |
| FCBI | 34.83 | 32.71 | 0.73 |
| ICBI | 33.14 | 32.93 | 0.75 |
| MLP_gen | 18.03 | 35.57 | 0.87 |
| MLP_sp | **16.91** | **35.85** | **0.89** |

For each table, the leftmost column is the methods used to test the particular image, while the second column displays the MSE results, follows by PSNR and SSIM in the last column. As for SSIM, the measurement is in range of 0 to 1, the results closer to 1 represents higher resemblance of output result to original reference image.

## 6 Discussion

### 6.1 Comparison of different up-scaling method results

The results for nearest neighbour, bicubic interpolation, Chen et. al., NEDI, FCBI and ICBI are calculated using the images provided by Giachetti and Asuni [2], while the results for MLP_gen and MLP_sp are the results from the trained network. The simulation results show that MLP generally perform better than ordinary image up-scaling methods. In most of the cases, MLP yields very low MSE in contrast with other methods, while PSNR are about the same. SSIM index shows significant resemblance of original reference images and the up-scaled images from MLP.

Figure below shows the output images from different methods, zoomed and cropped into smaller region:

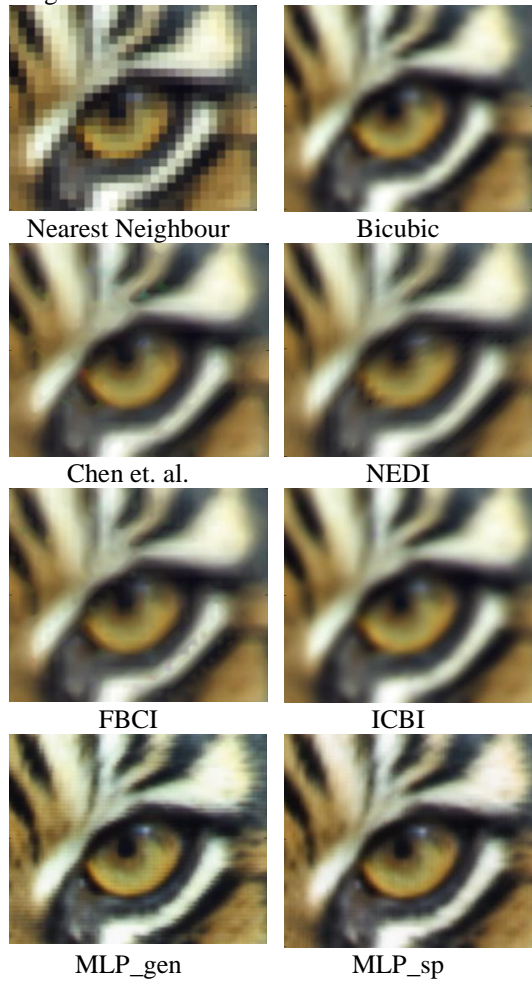

| Nearest Neighbour | Bicubic |
| Chen et. al. | NEDI |
| FBCI | ICBI |
| MLP_gen | MLP_sp |

**Fig. 3.** Comparison between different types of image up-scaling methods

Image up-scaled using Nearest Neighbour gives very jagged edges, losing much of the detail on tiger's fur. Bicubic interpolation, however gives better result but tends to over smooth the image, resulting in blurry image. NEDI however provides better edges but still facing the blurring effect. Meanwhile, method proposed by Chen et. al. and FCBI introduce unwanted artefacts to the image, and its solved by using the ICBI method proposed by Giachetti and Asuni. The results obtained from MLP, provides a clearer image without introducing artefacts and jagged edges, while maintaining the details on tiger's fur.

### 6.2 Comparison between MLP trained with different samples

Comparing both the MLP trained with different samples provided, generally the MLP trained with samples from different categories gives a more stable results for all the test cases, while MLP trained with samples from same the category performs significantly better when provided with adequate training samples.

The figures below show the comparison of MLP trained with different samples:

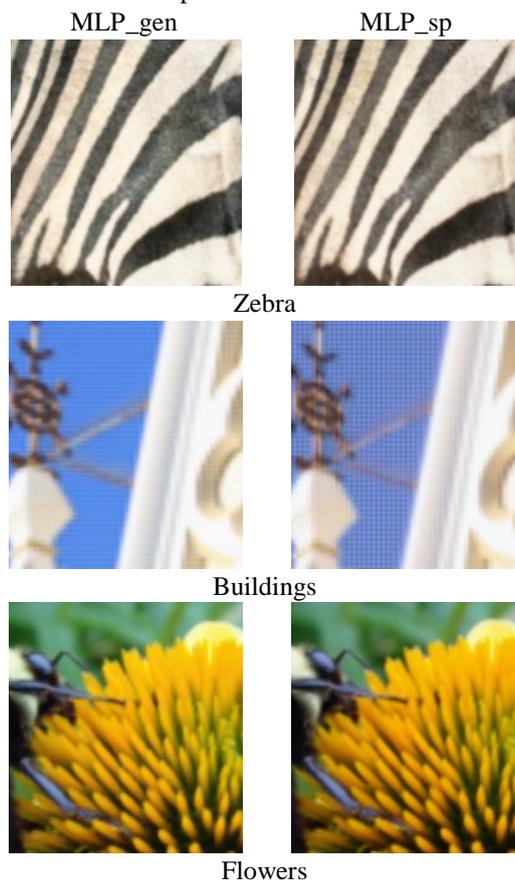

**Fig. 4.** Comparison of the MLP results trained using different samples

When provided with sufficient samples, MLP_sp trained with similar images can perform better than MLP_gen trained with samples from different categories. However, MLP_sp is not as robust as MLP_gen when handling colours beyond its training samples provided. For instance, MLP_sp trained with samples from building images unable to provide accurate result for sky colour, in which unwanted artefacts occurs in the sky region, while MLP_gen trained with various samples tackles this problem accurately.

### 6.3  Difference of original reference image and up-scaled image

Figures below show the difference between the original reference image and the up-scaled images from MLP_gen & MLP_sp:

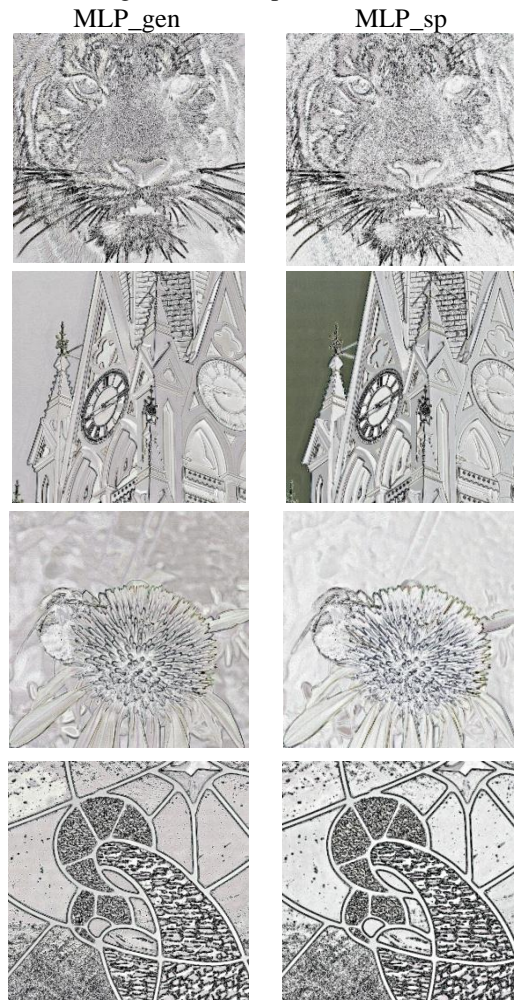

**Fig. 5.** The difference of original reference image and up-scaled images

Referring to the difference of images obtained in Figure 5, the colour of the images is inversed and scaled so comparison can be easily made. The brighter region represents the area where up-scaled image closely resembles the reference image, while darker colour means there is noticeable difference between the two images.

MLP performs reasonably well when provided with adequate training samples. The advantage of a neural network based image up-scaling method is the flexibility and adaptive nature of the network, which enable it to provide accurate results when the network is well trained. However, drawbacks exist in neural network as well. Training samples greatly affects the accuracy of the results, hence choosing the proper training samples plays an important role in judging the performance of a neural network based image up-scaling method.

In addition, MLP trained with single type of samples can accurately provide the up-scaled image, provided it's trained with sufficient training data. However, when there is some texture or colour which does not appear in training samples, the system will output the wrong results and creates unwanted artefacts. For instance, MLP_sp trained with building structures will have performance issue when there is part of the sky being exposed in the test image.

By comparing the two MLPs trained with different samples, MLP can upscale the image provided accurately for most of the regions, but suffers performance loss at most of the edges in the image.

Referring to the subjective test organised by Giachetti and Asuni [2], they requested a group of 12 people to compare and choose the best quality from different image up-scaling methods, and the average qualitative scores obtained shows that ICBI yields the best result. According to the quantitative results obtained previously, we believe that a similar subjective test perform on the MLP up-scaled images will provide higher rating compare to ICBI and other methods.

## 7    Conclusions & Future Works

To sum it all, a neural network trained with single type of samples performs splendidly on images of similar type, however it's not robust against images beyond the training samples or category. Meanwhile, a more robust neural network, trained with samples from various categories, can handle different images accurately without introducing artefacts. The adaptation ability is a major breakthrough and advantage of neural network, ought to be included in the future work of the research by fine tuning the trained neural network, allowing it to perform even better by adapting towards the current requirement. In addition, the training process can be sped up using Graphical Processing Unit (GPU) instead of CPU, hence reducing power consumption and training time without worrying about performance loss.

## Acknowledgment

The authors would like to present utmost gratitude to A. Giachetti and N. Asuni for their aid and guidance.

# References

1. Asuni, N., Giachetti, A.: Accuracy Improvements and Artifacts Removal in Edge Based Image Interpolation. Proc. 3rd Int. Conf. Comput. Vis. Theory Appl (VISAPP), 58-65 (2008)
2. Giachetti, A., Asuni, N.: Real-Time Artifact-Free Image Upscaling. IEEE Transactions on Image Proccessing 20(10), 2760-2768 (October 2011)
3. Tsai, R. Y., Huang, T. S.: Multiframe Image Restoration and Registration. In : JAI Press Inc., London (1984)
4. Tian, J., Ma, K.-K.: A Survey on Super-Resolution Imaging. Springer-Verlag London (2011)
5. Torrieri, D., Bakhru, K.: Neural Network Superresolution. IEEE, 1594-1598 (1997)
6. Egmont-Petersen, M., Ridder, D., Handels, H.: Image Processing with Neural Networks - A Review. Pattern Recognition 35, 2279-2301 (2002)
7. Sherrod, P. In: DTREG. Available at: http://www.dtreg.com/mlfn.htm
8. Chen, M. J., Huang, C. H., Lee, W. L.: A Fast Edge-Oriented Algorithm for Digital Images. Image Vis. Comput. 20, 805-812 (2002)
9. Elad, M., Fueur, A.: Restoration of a SIngle Super-Resolution Image from Several BLurred, Noise and Undersampled Measured Images. IEEE Transactions on Image Processing 6(12), 1646-1658 (1997)
10. Freeman, W. T., Jones, T. R., Pasztor, E. C.: Example-based Super-Resolution. IEEE Computer Graphics and Applications 22(2), 56-65 (2002)
11. Sudheer Babu, R., Sreenivasa Murthy, K. E.: A Survey on the Methods of Super-Resolution Image Reconstruction. International Journal of Computer Application 15(2) (February 2011)
12. Liu, H. Y., Zhang, Y. S., Ji, S.: Study in The Methods of Super-Resolution Image Reconstruction. The International Archives of the Photogrammetry, Remote Sensing and Spatial Information Sciences XXXVII(B2) (2008)
13. Baker, S., Kanade, T.: Limits on Super Resolution and How to Break Them. IEEE TPAMI 24(9), 1167-1183 (2002)